\documentclass[10pt,twocolumn,letterpaper]{article}

\usepackage{cvpr}
\usepackage{times}
\usepackage{epsfig}
\usepackage{graphicx}
\usepackage{amsmath}
\usepackage[dvipsnames]{xcolor}
\usepackage{multirow}
\usepackage{amssymb}

\usepackage[pagebackref=true,breaklinks=true,letterpaper=true,colorlinks,bookmarks=false]{hyperref}

\cvprfinalcopy 

\def\cvprPaperID{10} 

\ifcvprfinal\fi
\begin{document}

\title{Deep Anchored Convolutional Neural Networks}
\author{Jiahui Huang \; \; \; Kshitij Dwivedi \; \; \; Gemma Roig\\
Singapore University of Technology and Design\\
{\tt\small jiahui\_huang@sutd.edu.sg}, {\tt\small kshitij\_dwivedi@mymail.sutd.edu.sg}, {\tt\small gemma\_roig@sutd.edu.sg}
}

\maketitle
\thispagestyle{empty}

\begin{abstract}
Convolutional Neural Networks (CNNs) have been proven to be extremely successful at  solving computer vision tasks. State-of-the-art methods favor such deep network architectures for its accuracy performance, with the cost of having  massive number of parameters and high weights redundancy. Previous works have studied how to prune such CNNs weights.  In this paper, we go to another extreme and analyze the performance of a network stacked with a single convolution kernel across layers, as well as other weights sharing techniques. We name it Deep Anchored Convolutional Neural Network (DACNN). 
Sharing the same kernel weights across layers allows to reduce the model size tremendously, more precisely, the network is compressed in memory by a factor of $L$, where $L$ is the desired depth of the network, disregarding the fully connected layer for prediction. The number of parameters in DACNN barely increases as the network grows deeper, which allows us to build deep DACNNs without any concern about memory costs. 
We also introduce a partial shared weights network (DACNN-mix) as well as an easy-plug-in module, coined regulators, to boost the performance of our architecture. 
We validated our idea on 3 datasets: CIFAR-10, CIFAR-100 and SVHN. Our results show that we can save massive amounts of memory with our model, while maintaining a high accuracy performance.

\end{abstract}
\begin{figure}[t]
\label{figure1}
\begin{center}
   \includegraphics[width=1\linewidth]{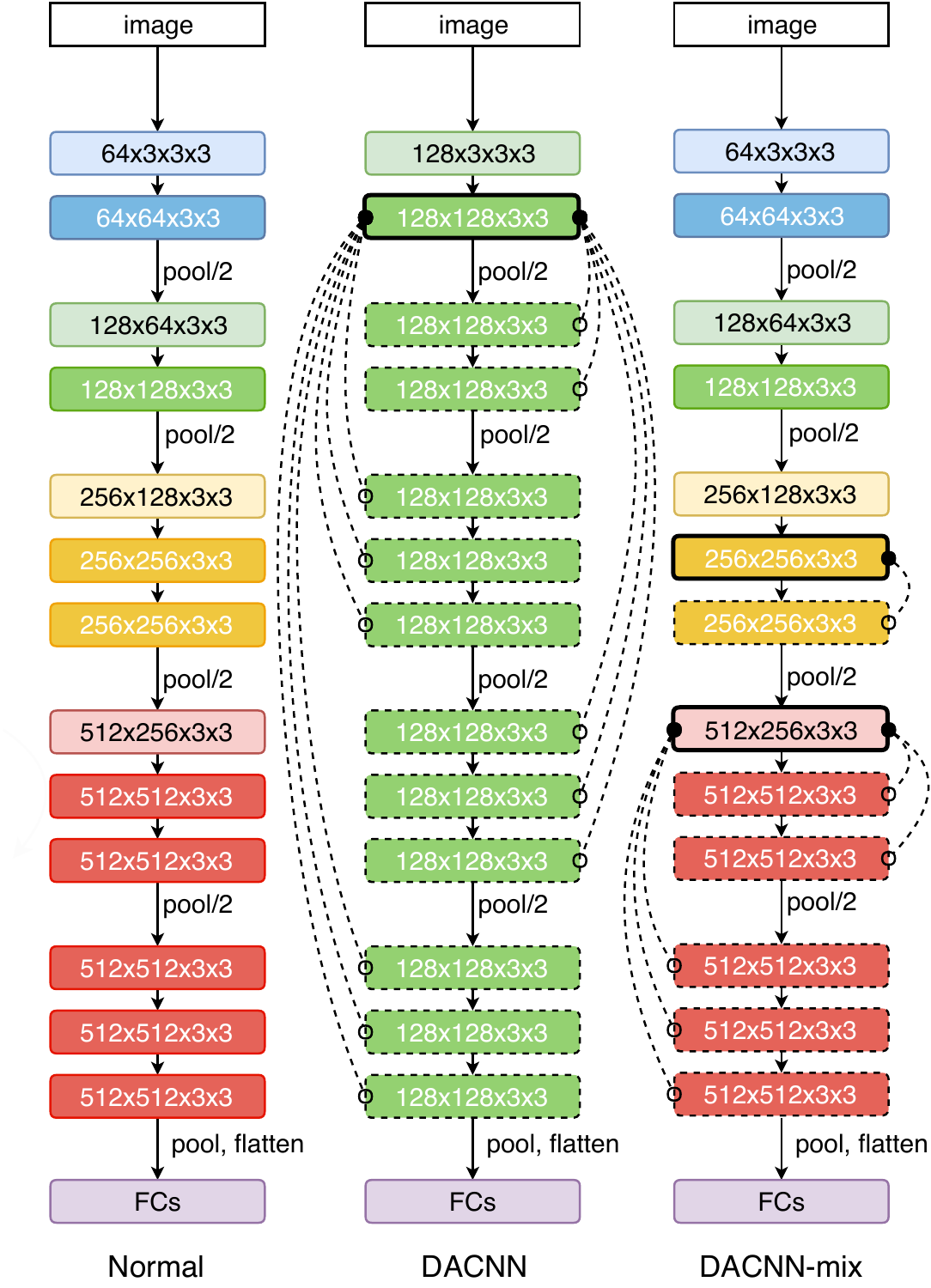}
\end{center}
   \caption{Example of a network with 13 convolutional layers. Kernels connected by dash lines represents they share the same set of weights, color codes denotes different kernel dimensions. {\bf Left: }Normal Architecture (14.9 million parameters) as reference. {\bf Middle: }Plain DACNN Architecture (0.16 million parameters). {\bf Right: }Mixed DACNN (4.73 million parameters). For more detailed model comparison, please refer to Table \hyperref[table6]{6}.}
\end{figure}

\section{Introduction}
Since the famous AlexNet~\cite{alexnet} outperformed all its competitors on ILSVRC2012 challenge~\cite{imagenet} in 2012, Convolutional Neural Networks (CNNs)  dominated almost all other approaches in computer vision tasks in the past 6 years~\cite{inception,localization,vgg,inception2}.  He \etal\cite{ResNet} proposed Residual Networks,  which allows to build extremely deep CNNs while still keep them optimizeable. Since then, the general trend is that CNNs have grown deeper and wider to achieve better performance~\cite{DPN,wideResNet,DeepNW}. As a result, current deep networks often come with vast amount of parameters and highly redundant weights~\cite{redundant}, while the performance gained is very limited compared to the number of parameters increased. For instance, ResNet-101 has only 1.1\% gain in accuracy compared to ResNet-50~\cite{ResNet} on ImageNet classification task, while the number of parameters almost doubled. This unbalance between model size increment and performance boost has become a severe problem yet to be tackled.

To compress CNNs, network weights pruning techniques have been introduced to remove some of the unnecessary filters~\cite{deepcompression,Han,pruningsoftware,prune_agent,handprune,data_free_pruning,compactCNN}. The aim of such techniques is to have smaller models without compromising the accuracy performance. Hao \etal\cite{handprune} proposed to prune weights according to their summed absolute weights. Huang \etal\cite{prune_agent} introduced a pruning agent to help analyze which filters are to be removed. These pruning methods adapt a ``subtraction" fashion to reduce the number of parameters, which means that once the original architecture is set, the performance of the model can barely increase since they consist on cutting off filters from it. If one needs further improvements on the model, the only way is to rebuild the original architecture and re-run the pruning algorithm again. Moreover, as far as we know, these techniques are not inbuilt in existing deep learning libraries and hence not widely adopted in most of the applications.

Our work focuses on addressing network memory compression problem in an ``addition" fashion. First, we propose a novel architecture that stacks a single convolution kernel over all layers (Figure \hyperref[figure1]{1}, middle), and we  extend it to partially share weights between pooling layers (Figure \hyperref[figure1]{1}, right). We call it \textit{deep-anchored-convolutional neural network (DACNN)}. We also introduce an ``easy-plug-in" way to add few extra parameters into the DACNN base model to boost the performance. Because the number of extra parameters introduced is determined by the model designer, this method provides an easy control of the trade-off between model size and model performance. In addition, the idea is easily realizable in code and it is applicable to most of the existing architectures.

We provide a detailed analysis on different DACNN architectures, as well as discuss how to efficiently add extra parameters to DACNN to achieve better performance with high memory compression rates. We demonstrate the efficiency and efficacy of our proposed method on CIFAR-10, CIFAR-100~\cite{cifar} and SVHN~\cite{svhn} datasets.   

\section{Related Works}
In this section we revisit network pruning techniques, as well as specific network models, namely ShaResNet~\cite{shaResNet}, SqueezeNet~\cite{squeezenet} and residual adapters~\cite{residualadapter}, which are related to our work.

\paragraph{Network Pruning.}
Over the past years, network pruning has become a  popular topic to compress the model size of neural networks. Han \etal\cite{Han} developed a method that replaces weights below a threshold with zeros. It forms a sparse matrix with less parameters, and then trains it for several iterations to achieve promising compression versus accuracy results. They further introduced quantization and huffman encoding into their Deep Compression~\cite{deepcompression} pruning method.  Huang \etal\cite{prune_agent} proposed a data-driven pruning method by introducing a pruning agent to remove unnecessary CNN filters. They use reinforcement learning to train the agent to prune the network while retains the network with a desired performance. Many of these pruning methods require backbone framework modification of the model, which reduce their applicability. Some of these methods even require dedicated hardware support~\cite{pruningsoftware}.  As a result, network pruning methods are not adopted on most of the existing DNN architectures. In addition, due to the nature of pruning, performance of the network can barely increase, and reconstruction of the base model and re-running of the pruning algorithm are required if one wants to improve the network performance.

\paragraph{ShaResNet.}
Boulch~\cite{shaResNet} proposed sharing weights among residual blocks to reduce the number of parameters without losing much performance. More concretely, a basic residual block~\cite{ResNet} is composed of 2 convolution operations with filter size $3\times 3$, ShaResNet uses shared weights to replace all the second convolution kernel within blocks that operate in the same spatial resolution (between 2 pooling layers). Thus, nearly half of the parameters from convolution can be cut off. A similar technique is applied to bottleneck blocks in deeper ResNets. Despite achieving promising results, this method is not flexible, as only one convolution is shared across blocks, and it's difficult to cut more parameters or increase performance from this architecture.

\paragraph{SqueezeNet.}
Iandola \etal\cite{squeezenet} introduced SqueezeNet, which consist on  replacing some of the $3\times 3$ convolution filters with $1\times 1$ filters as well as reducing the number of input channels to $3\times 3$ filters. They also pushed the downsampling of activation maps towards the end of the architecture to improve accuracy. With this approach, they were able to achieve same performance compared to AlexNet~\cite{alexnet} while using 50$\times$ less parameters. Yet, their method is not applicable to very deep networks such as ResNets~\cite{ResNet}, because it requires carefully designed structure for every layer, and downsampling in later layers in deep networks greatly increases computational cost.

\paragraph{Residual Adapters.}
Residual adapters were introduced by Rebuffi \etal\cite{residualadapter} as a technique for multi-task learning. They plug task-specific residual adapter modules (banks of $1\times 1$ convolution kernels) into residual blocks of the network. For different task domains, only these adapters varies while the rest parameters (90\%) remains the same. Since these $1\times 1$ convolution kernels are relatively small in size and they helps to regulate convolutional layer expressions, we introduce them as extra parameters to DACNNs to increase performance efficiently. In our work we call these $1\times 1$ convolution kernels \textit{regulators}.

\section{Anchored Weights Convolution}
Here, we introduce to notation that we use in the rest of the paper, as well as the components of our proposed anchored weights convolution architecture.

\paragraph{Notation.} Let us consider $X_0$ as an input image to the convolutional network with $L$ number of layers. $F_l(\cdot)$ denotes the transformation of each layer $l$, which can be a combination of convolution~\cite{CNN} (weights represented by $W^{conv}_l$), batch normalization~\cite{BN} (weights represented by $W^{bn}_l$), non-linear activation (ReLU~\cite{relu} in our case) or pooling~\cite{pooling} . We refer the output of layer $l$ as $X_l$.

\begin{figure}[t]
\phantomsection
\label{struc1}
\begin{center}
   \includegraphics[width=0.5\linewidth]{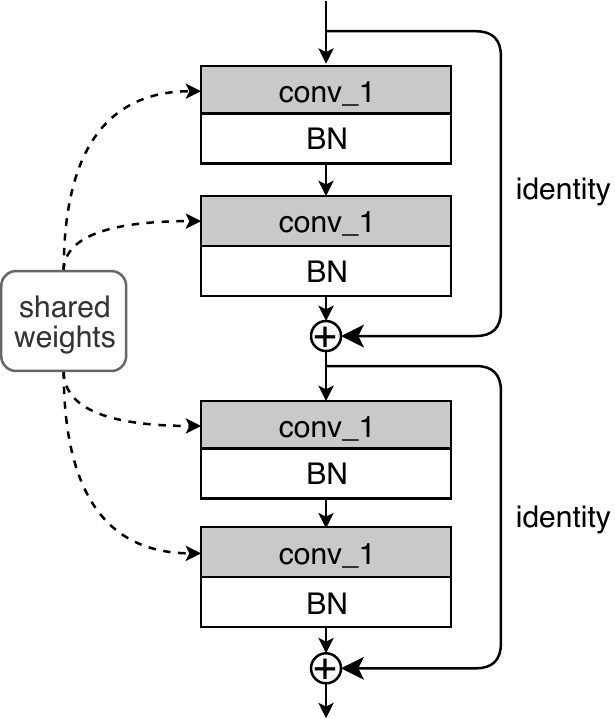}
\end{center}
   \caption{DACNN base block with residual learning: parameters for convolution are shared in all layers, while parameters for Batch Normalization are free.}
\end{figure}

\paragraph{Weights Sharing.}
In most of the CNN architectures, transformation function $F_l(\cdot)$  have different parameters for each layer: $W^{conv}_l$ and $W^{bn}_l$. A transformation of each layer can be represented as:
\begin{align}
    X_l = F_l(X_{l-1}, \{W^{conv}_l, W^{bn}_l\})
\end{align}
In our DACNN architecture (figure \hyperref[figure1]{1}), we first set a constant $C$ as the number of activation map channels for all layers, then we use one layer of transformation to expand a 3-channel-image to a desired activation map with $C$ channels: 
\begin{align}
    X_1 = F_1(X_0, \{W^{conv}_1, W^{bn}_1\})
\end{align}
From the second layer on wards, we internalize a set of global convolution weights $W^{conv}_G$ of shape $(C \times C \times filter size)$, and those are applied to every transformation function throughout the entire network:
\begin{align}
    X_l = F_l(X_{l-1}, \{W^{conv}_G, W^{bn}_l\})\quad (for\, l \geq2)
\end{align}
In this way, only weights for the first convolution $W^{conv}_1$, weights for global convolution $W^{conv}_G$ and weights for each layer's batch normalization $W^{bn}_l$ need to be initialized and trained,  greatly reducing the total number of parameters.

\paragraph {Batch Normalization.}
Batch normalization~\cite{BN} was first introduced as a technique to improve the performance and stability of deep neural networks. As we will show in the sequel, it is a crucial component in our architecture for achieving a good accuracy performance, as it allows scaling the activation map $A_l$ of each layer:
\begin{align}
    X_l = W^{bn}_l A_l.
\end{align}
Since the parameters of batch normalization for each layer are different, we can obtain different transformation functions across layers, and thus distinguish our work from simply stacking convolution kernels. In section \hyperref[sec:sec4]{4}, we  further discuss that scaling with batch normalization is a crucial operation in our architecture for performance. 

\paragraph {Residual Learning.}
ResNet~\cite{ResNet} is an architecture composed of residual blocks. Each block's  output is an  element-wise addition of input and activation:
\begin{align}
    X_l = X_{l-1}+F_l(X_{l-1}, \{W^{conv}_l, W^{bn}_l\}).
\end{align}
In our case, we adapt the idea of residual learning by using residual blocks with shared convolution weights (Figure \hyperref[struc1]{2}):
\begin{align}
    X_l = X_{l-1}+F_l(X_{l-1}, \{W^{conv}_G, W^{bn}_l\})
\end{align}
In this way, we are able to increase the depth of the network keeping it optimizable. In our case, for all DACNN architectures that are deeper than 17 layers, residual learning is applied.

\begin{figure}[t]
\phantomsection
\label{struc2}
\begin{center}
   \includegraphics[width=1\linewidth]{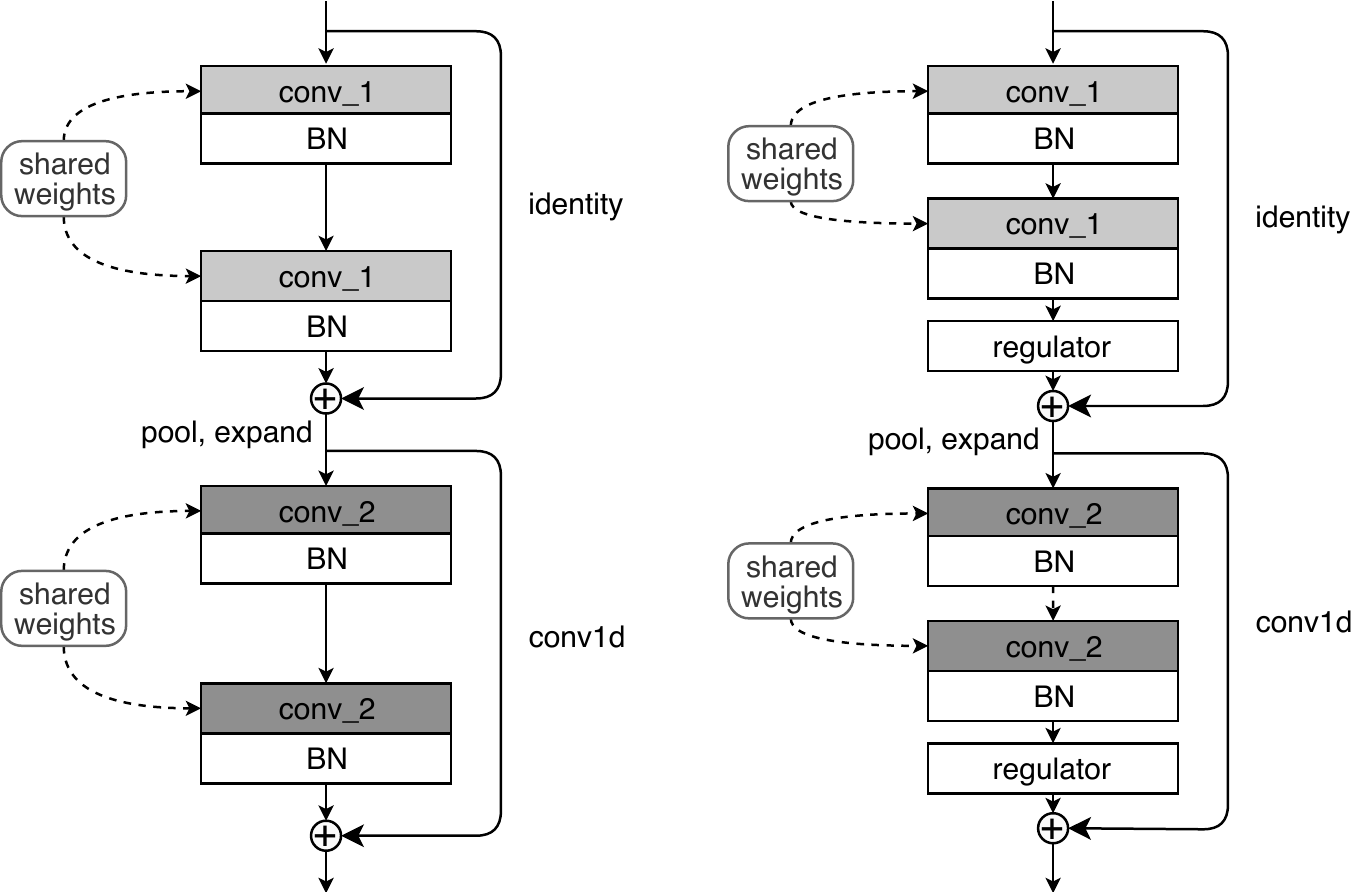}
\end{center}
   \caption{{\bf Left:} Base block for mixed DACNN, parameters are only shared between blocks operates in the same spatial resolution (between pooling layers). {\bf Right: }Base block for mixed DACNN with regulators, one regulator is appended to each block with free parameters.}
\end{figure}

\paragraph{Mixed Architecture.}
Since convolution kernels may behave differently when receptive field changes, we also adapted our approach of sharing weights on layers that only operates at same spatial resolution. In other words, instead of sharing one convolution weights throughout the entire network, we separate the network into sections by pooling~\cite{pooling} layers, and weights will only be shared within each section (Figure \hyperref[struc2]{3}, left). In this way, number of channels can be expanded as the network goes deeper. As a trade-off, more parameters will be needed:
\begin{itemize}
    \item[-] One transition layer for each channel expansion (one more $1\times 1$ convolutional layer needed if residual learning is adopted).
    \item[-] One convolutional layer for each section (as shared weights).
    \item[-] Batch normalization~\cite{BN} weights for all layers.
\end{itemize}
For example, if the number of channels expands 4 times in an architecture: $(3\rightarrow64\rightarrow128\rightarrow256\rightarrow512)$ , 8 sets of convolution weights need to be initialized: 4 for expansion and 4 for section weights sharing.

\paragraph{Regulators.}
We also provide an easy-plug-in way to improve the performance of DACNNs, which we call regulators. A single regulator is constructed with a $1\times 1$ convolution kernel~\cite{CNN}, a batch normalization~\cite{BN} layer and a ReLU~\cite{relu} activation layer, as illustrated in Figure \hyperref[struc2]{3}, right. All parameters in a regulator are not shared and it can be plugged in anywhere of the network as long as dimension matches, it helps to regulate the output of each convolution layer with shared weights. In deep architectures that adapt residual learning, we argue that 1 regulator for each residual block is enough to achieve a desirable performance. We will also provide detailed experiment results on how many and where to add these regulators in the later section.

\paragraph{Implementation details.} For plain DACNNs, we use a $3\times 128\times3\times 3$ convolution kernel to expand a 3-channel image to a 128-channel activation map, and then followed by a $128\times 128\times 3\times 3$ kernel stacked $L$ times, where $L$ is the desired depth of the network, in addition, all convolution kernels above are followed by a batch normalization~\cite{BN} layer (free parameter) and a ReLU~\cite{relu} activation layer. For DACNNs that are deeper than 17 layers, residual learning is adopted to keep them optimizable. For mixed DACNNs, $(3\rightarrow64\rightarrow128\rightarrow256\rightarrow512)$ is adopted as channel expansion pattern for all architectures.

\section{Experiments}
\label{sec:sec4}
In this section, we  provide detailed analysis and results of DACNN on different perspectives. Experiments are conducted on Cifar 10, Cifar 100~\cite{cifar} and SVHN~\cite{svhn} datasets.
\subsection{DACNN Analysis}
We conducted thorough experiments on CIFAR-100 dataset~\cite{cifar} to evaluate our DACNN. We analyze the importance of batch normalization, the role of the depth of the architecture with respect to the number of parameters, as well as all the proposed additional components described in the previous section. 

\paragraph{Dataset and training settings.} We use CIFAR-100 dataset~\cite{cifar} to perform the analysis. The dataset contains 60,000 32x32 color images in 100 different classes (600 images per class). It is split into training set and test set with the ratio of 5:1. All models in this section are trained 90 epochs with random horizontal flip as data augmentation~\cite{alexnet}, for preprocessing, we normalize the data using the channel means and standard deviations as in~\cite{alexnet}. The networks are updated with ADAGRAD~\cite{adagrad} optimizer with learning rate set to 0.1 and decreases to 0.01 from 45th epoch on wards.

\paragraph{Importance of Batch Normalization.}
Here we analyze the impact of batch normalization (BN)~\cite{BN} in our DACNN architectures. We trained 2 models on CIFAR-100 dataset: a 14-layer plain DACNN (VGG~\cite{vgg} based) and a 18-layer plain DACNN with residual learning (ResNet~\cite{ResNet} based), they are trained both with and without BN for each layer. As we show in the results in Table \hyperref[table1]{1}, the network performs poorly without batch normalization in both tested models, giving an error of almost random guessing. Our hypothesis of this behavior is that BN helps scale the output feature map after every shared filter, thus introduces some divergence in to the network rather than simply stacking weights.  Therefore, for the rest of the experiments, all DACNN configurations are equipped with free batch norm parameters as a default setting.
\begin{table}
\phantomsection
\label{table1}
\begin{center}
\begin{tabular}{l| c | c}
\hline
 & w.o  BN &  w BN \\
\hline
DACNN14 (VGG based) & 97.48 & 42.29 \\
\hline
DACNN18 (ResNet based) & 97.17 & 38.63 \\
\hline
\end{tabular}
\end{center}
\caption{TOP-1 error (\%) on CIFAR-100 dataset. We compare DACNN network without batch normalization (BN), and DACNN with free BN weights for every layer. VGG and ResNet are selected as base template for 14 and 18 layer DACNN, respectively.  Both DACNN architectures have 128 as the fixed number of channels. The results shows that batch norm is crucial to our architecture.}
\end{table}

\begin{table}
\phantomsection
\label{table2}
\begin{center}
\begin{tabular}{c|ll}
\hline
\# layers & TOP-1 err. (\%) & \multicolumn{1}{l }{\# param (M)} \\ \hline
3         & 56.91 (54.41)   & 0.164 (0.45)                      \\
5         & 40.11 (33.74)   & 0.164 (0.74)                      \\
7         & 42.22 (32.66)   & 0.164 (1.03)                      \\
9         & 42.40 (32.67)    & 0.164 (1.33)                      \\
11        & 42.37 (30.36)   & 0.165 (1.63)                      \\
14        & 42.29 (30.44)    & 0.165 (2.06)                      \\ \hline \hline
18        & 38.63 (28.31)    & 0.166 (2.66)                      \\
34        & 38.88 (27.22)      & 0.168 (5.02)                     
\end{tabular}
\end{center}

\caption{TOP-1 test error (\%) on CIFAR-100: Layers vs \# parameters. Data inside parentheses are for architectures without sharing weights for comparison. All models here have 128 number of channels through out the network, for architectures deeper than 17 layers, residual learning is applied. For DACNNs, number of parameters barely increase as the network goes deeper. }
\end{table}

\paragraph{Impact of Depth in DACNN.}
Here, we provide comparison between DACNNs of different depths, we also compare them to networks without sharing the kernel weights with the same structures. As in the batch normalization experiment, all networks have 128 as fixed number of channels, pooling layers are inserted in between convolutions. In addition, for architectures deeper than 17 layers, residual learning is adopted to the entire network.

Results are shown in Table \hyperref[table2]{2}. The error rate of plain DACNN drops as the network goes deeper up to the 5-th layer. The next drop is when we introduce residual learning into DACNNs (see the 18-layer network). Compare to networks with free weights, model sizes of DACNNs do not increase much as the network grows. However, we observe that simply stacking plain DACNN kernels and increasing the depth barely benefit the performance of our architecture, we explore further improvements which are discussed in the sequel.

\begin{figure*}
\phantomsection
\label{figure4}
\begin{center}
\includegraphics[width=1\linewidth]{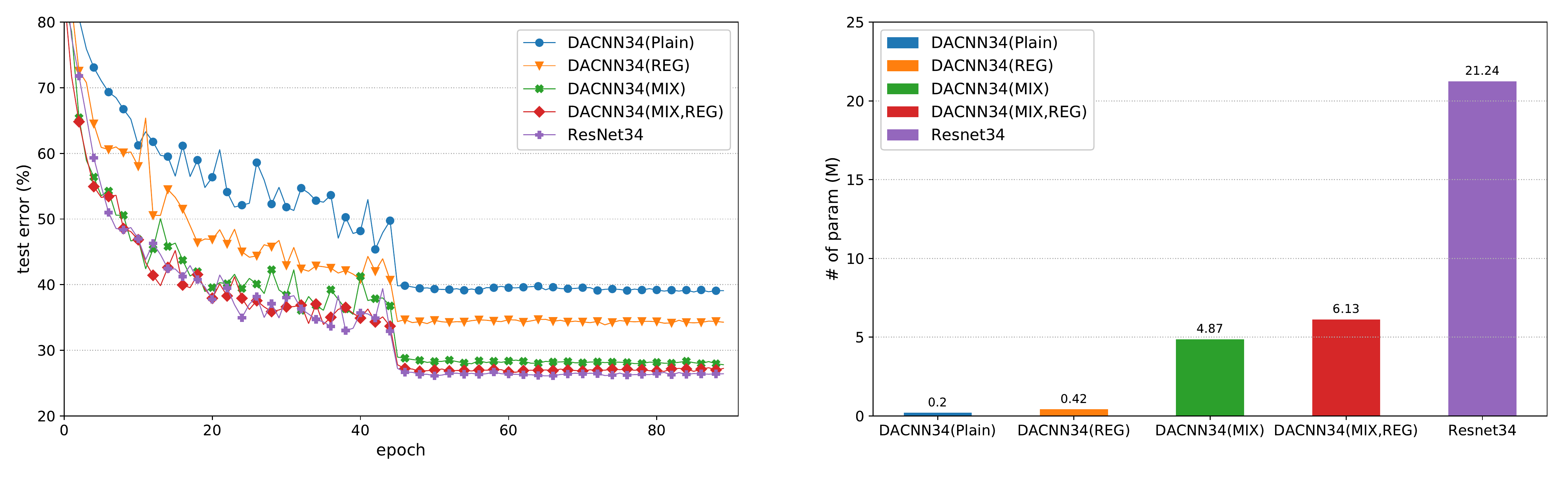}
\end{center}
   \caption{{\bf Left: }TOP-1 test error (\%) on CIFAR-100 dataset. Solid curves denote DACNN and it variations, dotted curve denotes ResNet-34 as comparison. {\bf Right: }Number of parameters of these architectures. Mixed DACNN-34 and mixed DACNN-34 with regulators achieved similar performance as ResNet34 while using much less parameters.}
\end{figure*}

\paragraph{Mixed DACNN.}
Next we evaluate mixed architectures for DACNNs. We choose VGG, ResNet18 and ResNet34~\cite{ResNet,vgg} as our base architectures. As we introduced earlier, mixed DACNN architectures require extra parameters whenever channel dimension expands, since all architectures above share the same channel expansion pattern, the number of parameters required are almost the same (for ResNet based architectures, one more $1\times 1$ convolution is needed for each shortcut expansion).

As shown in results of Table \hyperref[table3]{3}, with greatly reduced number of parameters, performance of mixed DACNNs are comparable to VGG and ResNets with same number of layers. Mixed DACNN18 has only 0.41\% performance drop compared to ResNet18, but model number of parameters is reduced by 55\%. Although increasing the depth doesn't benefit performance of mixed DACNNs, it doesn't increase the number of parameters neither. In later section, we argue that deeper DACNNs have higher capacity for improvements. 

\begin{table}
\begin{center}
\phantomsection
\label{table3}

\begin{tabular}{l| c | c}
\hline
Template & TOP-1 err. (\%) &  param (M) \\
\hline
VGG & 30.20 & 14.90\\
\hline
DACNN-14(MIX) & 30.54 & 4.73 \\
\hline
\hline
ResNet-18 & 27.31 & 11.13 \\
\hline
DACNN-18(MIX) & 27.72 & 4.90 \\
\hline
ResNet-34 & 26.10 & 21.24 \\
\hline
DACNN-34(MIX) & 27.70 & 4.91 \\
\hline
\end{tabular}
\end{center}

\caption{Classification results on CIFAR-100 dataset. We choose VGG, ResNet18, and ResNet34 as our base models. Mixed DACNNs obtained desirable performances with much less parameters compared to original VGG and ResNets.}
\end{table}

\label{table4}
\begin{table}
\begin{center}
\begin{tabular}{c| c | c}
\hline
SECTION & TOP-1 err. (\%) &  extra param \\
\hline
NONE & 38.63 & - \\
\hline
section 1 & 38.28 & 32K \\
\hline
section 2 & 38.09 & 32K \\
\hline
section 3 & {\color{blue}{\bf 37.46}} & 32K \\
\hline
section 4 & 38.07 & 32K \\
\hline
ALL & {\bf 35.39} & 128K \\
\hline
\end{tabular}
\end{center}

\caption{Allocations of regulators.Plain DACNN-18 was selected as the testing model, the network is cut into 4 sections by 3 pooling layers, and regulators are appended to residual blocks in each section separately. ALL denotes all sections, {\bf bold} denotes overall best result and {\color{blue}{\bf blue}} denotes best result on single section.}
\end{table}
\paragraph{Regulators.}
Here we examine the effectiveness of regulators ($1\times 1$ convolution kernels) in our DACNN architecture. First we test regulators on a plain 18 layer DACNN with residual learning, we separate the network into 4 sections by pooling layers, and experiment to add regulators to different sections separately (note that we only append one regulator into each residual block). Results are shown on Table \hyperref[table4]{4}. We observe that the performance of the network increases by a considerable margin as we append regulators to different sections. Since each regulator is a $1\times 1$ convolution kernel, only few extra parameters are added to the network. According to the results, appending regulators to the 3rd section gives the best efficiency, we dropped 1.17\% on the error rate compared to plain DACNN18 using only 32K parameters, and by appending regulators to all sections, we are able decrease the error rate by 3.24\%.

We also analyze the effect of  using  regulators to models with different depth. In this experiment,  regulators are appended to all sections to give better performance.  Results are shown on Table \hyperref[table5]{5}. As expected, deeper networks give better performance since they have more residual blocks to fit in regulators. On a 34-layer DACNN, we are able to obtain about 1.5\% drop on the error rate compare to a 18-layer DACNN with more regulators appended (0.34M more parameters).

\begin{table}
\phantomsection
\label{table5}
\begin{center}
\begin{tabular}{c| c | c}
\hline
model & TOP-1 err. (\%) &  extra param (M) \\
\hline
DACNN-18 (REG)& 35.39 & 0.32 \\
\hline
DACNN-24 (REG)& 34.86  & 0.52 \\
\hline
DACNNt-34 (REG)& {\bf 33.88} & 0.68 \\
\hline

\end{tabular}
\end{center}
\caption{The effect of regulators on plain DACNNs. Extra parameters are computed with respect to a plain DACNN-18. All networks are trained and tested on CIFAR-100 dataset. REG denotes regulators, {\bf bold} denotes our best result.}
\end{table}

Lastly, we combine everything above together. We apply both mixed structure and regulators to DACNNs, the results can be found at the bottom of Table \hyperref[table6]{6}. In comparison with ResNet-18, Mixed DACNN-34 with regulators obtains better accuracy while using only half number of the parameters; Comparing to ResNet-34, mixed DACNN-34 with regulators are 0.56\% lower in accuracy, but the model size is $4\times$ smaller.

\paragraph{Model Efficiency.}
Here, we evaluate model efficiency by considering model size and performance. Architectures with higher performance and less parameters will be considered as high efficiency models. 

\begin{figure}[t]

\begin{center}
   \includegraphics[width=1\linewidth]{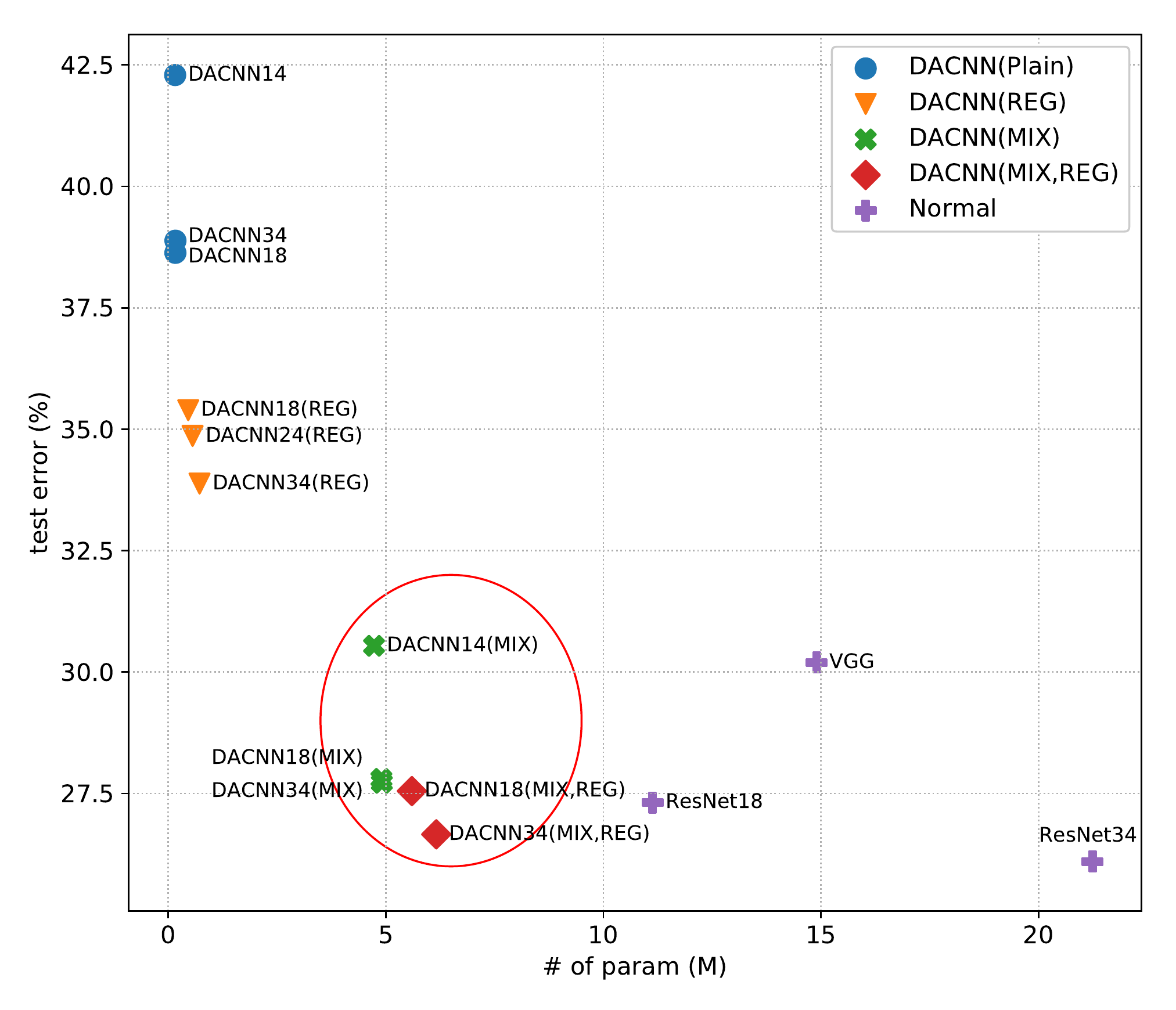}
\end{center}
\label{figure5}
   \caption{Model Efficiency. This is a plot of TOP-1 error rate on CIFAR-100 vs. number of parameters. Architectures with high model efficiency are at the bottom left of the graph. Mixed DACNNs and mixed DACNNs with regulators (circled with \textcolor{red}{red}) give highest model efficiency among all other models in our experiment.}
\end{figure}

We provide results of different architectures on CIFAR-100, as well as their number of parameters on Table \hyperref[table6]{6}. On Figure \hyperref[figure4]{4}, we plotted the testing curves and model sizes of 34-layer network architectures with different configurations. From the results, we observe that plain DACNNs are extremely small in model size but their accuracy are not competitive. Yet, as we introduce mixed structure and regulators into the model, the boost in accuracy is tremendous, while the number of parameters of resulting models are still smaller than VGG and ResNets~\cite{vgg,ResNet} by a large margin. For instance, mixed DACNN-34 with regulators has 15.1 million parameters fewer than ResNet-34. Figure \hyperref[figure5]{5} illustrates a plot of parameter efficiency, inwhich architectures with high model efficiency are expected  to be plotted on the bottom left of the graph. As shown on the figure, DACNNs with mixed structure and regulators are much  more efficient than plain DACNNs and normal architectures (VGG, ResNets~\cite{vgg,ResNet}).

\begin{table}[]
\begin{tabular}{l|cc}
\hline
method              & \multicolumn{1}{c}{TOP-1 err. (\%)} & \multicolumn{1}{c}{\# param (M)} \\ \hline
VGG (1-fc)          & 31.20                                 & 14.90                             \\
ResNet18            & 27.31                                & 11.13                            \\
ResNet34            & 26.10                                 & 21.24                            \\ \hline \hline
DACNN-14 (plain)      & 42.20                                 & 0.16                            \\
DACNN-18 (plain)      & 38.63                                & 0.16                            \\
DACNN-34 (plain)     & 38.88                               & 0.17                            \\ \hline \hline
DACNN-18 (REG)      & 35.39                                & 0.46                            \\
DACNN-24 (REG)      & 34.86                                 & 0.56                              \\
DACNN-34 (REG)      & 33.88                                 & 0.72                             \\ \hline \hline
DACNN-14 (MIX)      & 30.54                                & 4.73                             \\
DACNN-18 (MIX)      & 27.72                                 & 4.90                              \\
DACNN-34 (MIX)      & 27.80                                & 4.91                             \\ \hline \hline
DACNN-18 (MIX, REG) & 27.55                                & 5.60                              \\
DACNN-34 (MIX, REG) & {\bf 26.66}                               & {\bf 6.10}                           
\end{tabular}
\label{table6}
\caption{Classification results on {\bf CIFAR-100} dataset. All DACNN variations are shown on this table, together with VGG and ResNets for comparison. Accuracy of DACNN-34 is only a small margin lower than ResNet34, while cutting off 70\% of the parameters.}
\end{table}

\subsection{Classification Results on CIFAR-10 and SVHN}
To validate our method on other datasets, we also trained DACNNs on CIFAR-10 and SVHN~\cite{cifar,svhn}. CIFAR-10 has similar configuration as CIFAR-100 but with only 10 classes. SVHN is a house number recognition dataset obtained from Google Street View images, there are 73,257 images in its training set, and 26,032 images in the test set.

Mixed DACNN-18 with regulators and mixed DACNN-34 with regulators are selected in this experiment, we also trained ResNet-18 and ResNet-34 for comparision. On CIFAR-10, models are trained 90 epochs without data augmentation, learning rate was set to 0.1 and decreases to 0.01 at 45th epoch. On SVHN dataset, models are trained 60 epochs, following common practice,~\cite{maxout,DeepNW,svhn_cnn,networkinnetwork} no data augmentation is applied. Learning rate starts from 0.1, decreases as a factor of 10 for every 20 epochs , we use ADAGRAD~\cite{adagrad} as our optimizer in both cases. 

\begin{figure*}
\phantomsection
\label{figure6}
\begin{center}
\includegraphics[width=1\linewidth]{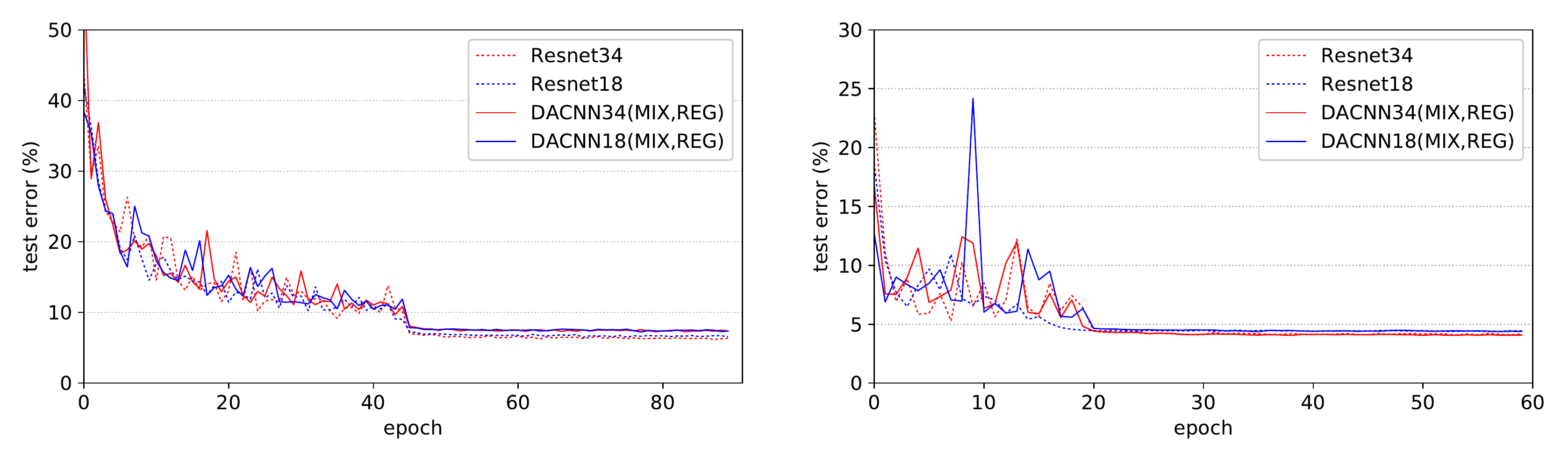}
\end{center}
   \caption{Traning \& Testing on {\bf CIFAR-10} and {\bf SVHN} dataset. Solid curves denotes results for DACNNs, and dashed curves denotes results for ResNets. We observe that ResNets are slightly better on CIFAR-10, while DACNNs are better on SVHN.}
\end{figure*}

The results are shown on Table \hyperref[table7]{7}, and Figure \hyperref[figure6]{6} is a plot of testing curves of both networks on both datasets. With this experiment, we validate the competitiveness and effective of our architecture as results are comparable to those with the architecture with free weights, and using much less number of parameters.

We also compare our method with 2 other pruning techniques: Agent Pruning by Huang \etal\cite{prune_agent}, and sparse matrix proposed by Han \etal\cite{Han}, results are shown on Table~\ref{table8}.
DACNNs are able to achieve high compression ratios with low accuracy drops compared to the other 2 methods. In addition, DACNNs are easier for implementation and deployment, while the rest two require data-driven fine tuning or additional software/hardware support.

\begin{table}[]
\begin{center}
\begin{tabular}{l|cc}
\hline
model             & CIFAR-10 & SVHN   \\ \hline
ResNet-18         & 6.47\%   & 4.38\% \\
ResNet-34         & 6.18\%   & 4.08\% \\ \hline
DACNN-18 (MIX,REG) & 7.26\%   & 4.37\% \\
DACNN-34 (MIX,REG) & 7.23\%   & 4.06\% \\ 
\end{tabular}
\end{center}
\label{table7}
\caption{TOP-1 errors on {\bf CIFAR-10} and {\bf SVHN} dataset. Results shows that DACNNs even outperformed ResNets on SVHN dataset.}
\end{table}

\begin{table}[h]
\begin{tabular}{l|cc}
\hline
method              & \multicolumn{1}{c}{Accuracy drop (\%)} & \multicolumn{1}{c}{Prune Ratio (\%)} \\ \hline
Agent Pruning          & 0.3                                & 27.1                             \\
Agent  Pruning         & 1.0                                & 37.0                             \\
Agent Pruning          & 1.7                                 & 67.9                             \\
SM   Pruning        & 1.3                                & 27.1                             \\
SM Pruning         & 6.9                                & 37.0                             \\
SM   Pruning        & 6.5                                 & 67.9                             \\
DACNN(M,R)            & {\bf0.79}                                & {\bf49.6}                            \\
DACNN(plain)            & {\bf10.2}                                & {\bf98.5}                            \\
\hline
\end{tabular}
\caption{Classification results on {\bf CIFAR-10} dataset. Agent Pruning \cite{prune_agent}, SM Pruning denotes sparse matrix Pruning \cite{Han} and (M,R) denotes Mixed Structure and Regulators respectively . ResNet 18 was selected as base architecture.}
\label{table8}
\end{table}

\subsection{Filter Visualization}
Here, we visualize the filters of DACNN. We apply similar technique as network fooling~\cite{fooling} for filter visualization: First input an image of random noise, then we optimize the input image with respect to each convolution layer using backpropagation~\cite{backpropagation}.

\begin{figure*}
\phantomsection
\label{figure7}
\begin{center}
\includegraphics[width=0.93\linewidth]{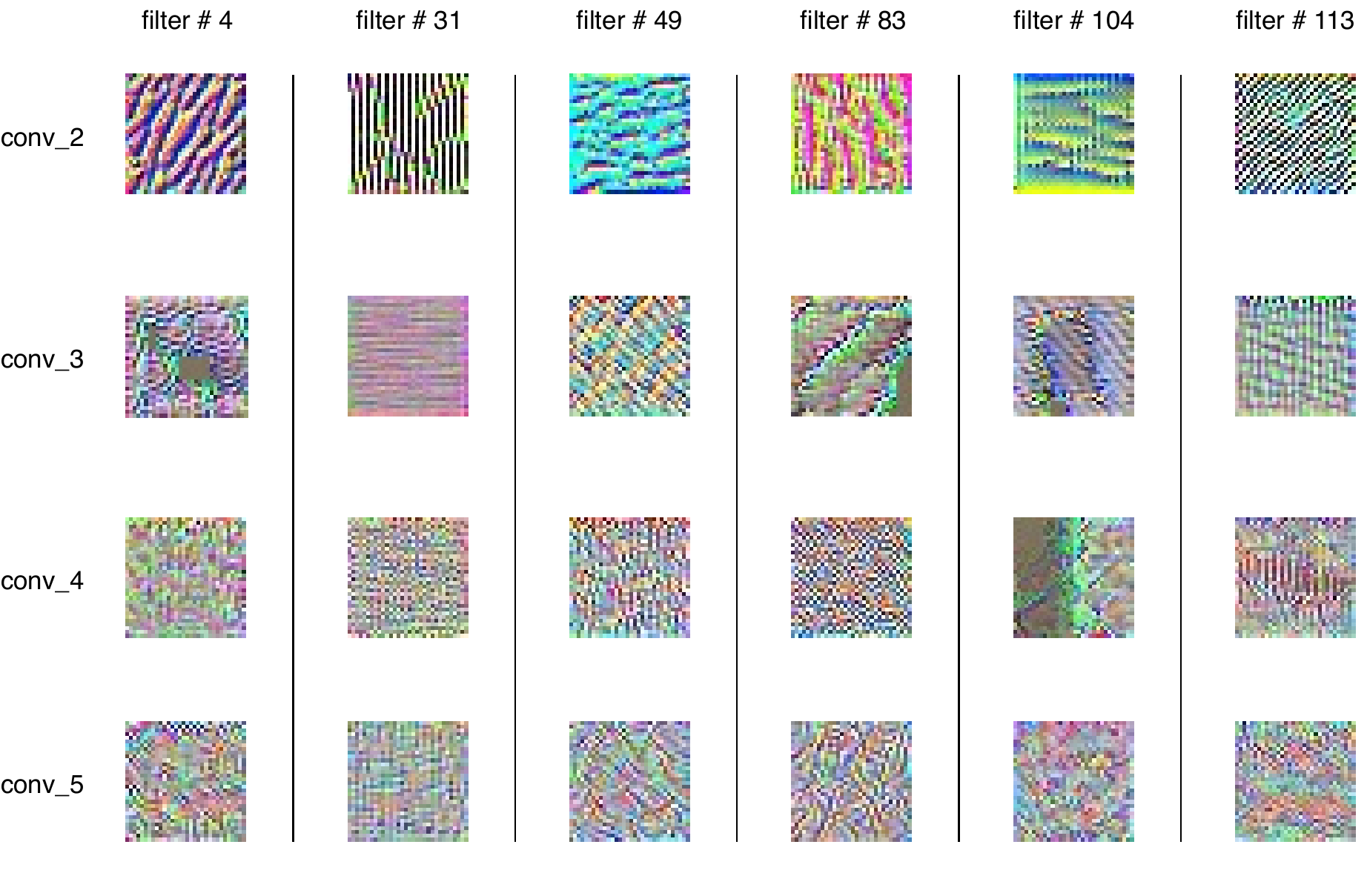}
\end{center}
   \caption{Filter Visualization. Each of the image above is optimized for the corresponding convolutional layer using backpropagation. We trained a 5 layer plain DACNN on CIFAR-10 dataset and selected 5 filters for demonstration.}
\end{figure*}

We train a 5 layer plain DACNN on CIFAR-10 ($32\times 32$) and plot the results of some filters on Figure \hyperref[figure7]{7}. The reason why we choose plain DACNN here, is that we want to demonstrate the output features can still be very diversified across layers even without implementations of regulators and mixed-architecture. This also illustrates the importance of  batch normalization to scale the filter responses.

\section{Conclusion}
We introduced a new convolutional neural network architecture, which we refer it as \textit{Deep Anchored Convolutional Neural Network (DACNN)}. It shares weights for convolution kernels across layers while keep parameters for Batch Normalization free. Due to high weights reusage, number of parameter of DACNN barely increases as the network goes deeper. Thus, it is a novel way for model size compression.

Since we observe that simply increasing the depth of DACNNs contributes little to the  performance, we also propose two ways to improve the performance of DACNNs: Mixed Structure and Regulators.

With these two methods, DACNN is an efficient model compression approach adopting an ``addition" fashion: First initialize a plain DACNN to a desired depth (deeper networks have higher capacity for further improvements). Then, selectively apply mixed structure and append regulators to achieve a desirable performance. As a result, DACNNs are able to obtain similar performance with much less parameters compare to some popular architectures like VGG and ResNet~\cite{vgg,ResNet}.

\paragraph{Acknowledgements.}
This work was funded by the SUTD-MIT IDC grant (IDG31800103). K.D. was also funded by SUTD Presidents Graduate Fellowship. 

{\small
\bibliographystyle{ieee}

\bibliography{ref}
}

\end{document}